# RADIOMIC SYNTHESIS USING DEEP CONVOLUTIONAL NEURAL NETWORKS

*Vishwa S. Parekh*[1,3], *Michael A. Jacobs*[1,2]

[1]The Russell H. Morgan Department of Radiology and Radiological Science and [2]Sidney Kimmel Comprehensive Cancer Center, The Johns Hopkins School of Medicine, Baltimore, MD 21205, USA
[3]Department of Computer Science, The Johns Hopkins University, Baltimore, MD 21208, USA

## ABSTRACT

Radiomics is a rapidly growing field that deals with modeling the textural information present in the different tissues of interest for clinical decision support. However, the process of generating radiomic images is computationally very expensive and could take substantial time per radiological image for certain higher order features, such as, *gray-level co-occurrence matrix(GLCM)*, even with high-end GPUs. To that end, we developed RadSynth, a deep convolutional neural network(CNN) model, to efficiently generate radiomic images. RadSynth was tested on a breast cancer patient cohort of twenty-four patients(ten benign, ten malignant and four normal) for computation of GLCM entropy images from post-contrast DCE-MRI. RadSynth produced excellent synthetic entropy images compared to traditional GLCM entropy images. The average percentage difference and correlation between the two techniques were $0.07 \pm 0.06$ and 0.97, respectively. In conclusion, RadSynth presents a new powerful tool for fast computation and visualization of the textural information present in the radiological images.

*Index Terms*— Radiomics, CNN, deep learning, radiomic synthesis.

## 1. INTRODUCTION

Radiomics is a rapidly advancing field that deals with extraction of quantitative textural and shape-based features from radiological images for advanced clinical decision support [1, 2]. The current techniques in radiomics extract features that model the intensity distribution and the inter-voxel relationships from a tissue of interest (e.g. tumor) [3-5]. In addition, these features could be used as imaging filters to produce radiomic images that could visualize and characterize the "radiological texture" across the complete radiological image [6]. However, the process of generating radiomic images is computationally very expensive, especially for the higher order statistical kernels. The time complexity for generating a GLCM entropy image for an $N \times N$ sized radiological image quantized to G gray levels using $W \times W$ sized statistical kernels is $O(N^2 G^2 W^2)$. Parallelization of the code for generating and executing GLCM radiomic images on a Tesla K40c (12GB, 2880 CUDA cores) takes approximately forty minutes using a $5 \times 5$ kernel on a $512 \times 512$ image.

To overcome this limitation, we developed a convolutional neural network (CNN) based deep learning model termed as RadSynth to efficiently generate radiomic images. CNNs are deep neural networks inspired by the hierarchical organization of the human cortex [7]. The shallow layers of the CNN capture local information while the deeper layers of the CNN capture more global information. The first layer of the CNN extracts edges and blobs which could be useful in identifying textural information present in the input image. Furthermore, the deeper CNN layers could potentially be trained to model more complex textural information in the input images.

In this paper, we present a texture synthesis deep CNN model, RadSynth for modeling the complex textural information present in radiological images and use that information to synthesize radiomic images. We illustrate the performance of RadSynth on a breast cancer dataset for extracting radiomic metrics from normal and lesion tissue.

## 2. MATERIALS AND METHODS

### 2.1. Clinical Data

We tested RadSynth on a retrospective breast cancer data consisting of twenty-four women. Of the twenty-four women, ten had malignant lesions, ten had benign lesions and four had no lesions. These patients were acquired in accordance with the Johns Hopkins University School of Medicine guidelines for clinical research under a protocol approved by the Institutional Review Board (IRB). All HIPAA agreements were followed for this study. We used the high spatial resolution post contrast dynamic contrast enhanced (DCE) magnetic resonance images (MRI) acquired for evaluating RadSynth. The imaging parameters for the post-contrast DCE-MRI were: $TR/TE = 5.67/2.9\ ms$, $field\ of\ view = 35cm \times 35\ cm$, $matrix\ size = 512 \times 512$, $Slice\ Thickness = 3\ mm$.

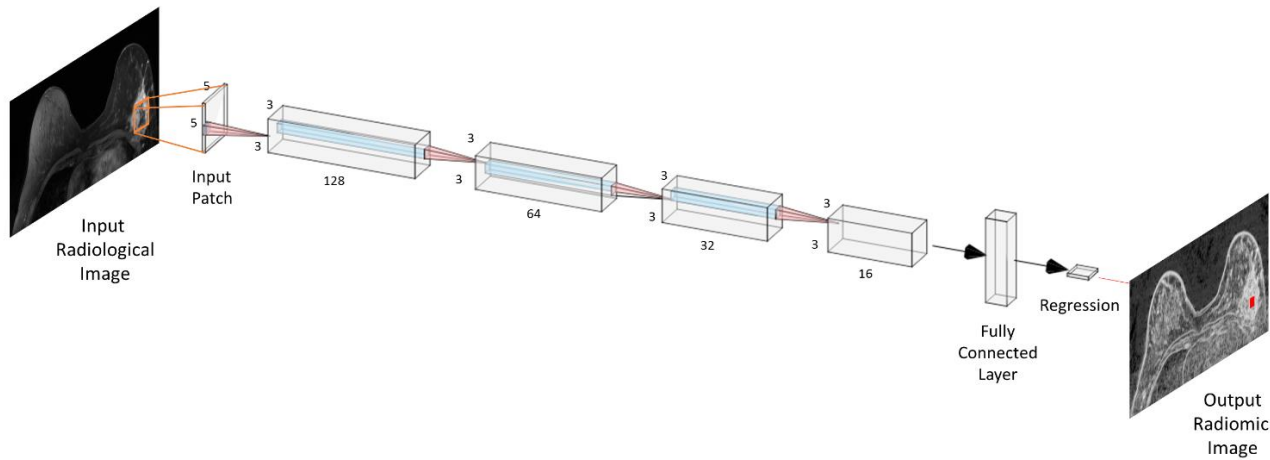

*Figure 1. Illustration of the RadSynth network architecture. RadSynth consisted of four layers with 128, 64, 32, and 16 filters respectively, followed by a fully connected layer and regression. Root-mean-squared error between the Haralick entropy and RadSynth GLCM entropy was implemented as the cost function for training RadSynth.*

### 2.2. Radiomics

The radiomic feature maps (RFMs) of GLCM based entropy were computed for each patient. The input parameters were set as follows:
Gray level quantization, $G = 64$
Size of the sliding window = $5 \times 5$
We term these as Haralick GLCM entropy images [4].

### 2.3. RadSynth

Training of RadSynth was done using image patches of size $5 \times 5$, consistent with the size of the sliding window used to generate GLCM entropy images. The network architecture for RadSynth is illustrated in **Figure 1**.

RadSynth consisted of four layers with 128, 64, 32, and 16 filters respectively, followed by a fully connected layer and a regression layer. Each layer of the 2D-CNN had the following components:
- Convolutional layer with trainable filters of size $3 \times 3$
- Batch Normalization
- ReLU activation function given by the following equation

$$f(x) = \begin{cases} 0, if\ x < 0 \\ x, if\ x \geq 0 \end{cases}$$

Max pooling layer with a $2 \times 2$ window was applied after the first and the second layer. The minibatch size was set at 2000 and a dropout of 0.2 was applied after the last convolutional layer.

RadSynth was tested using two-fold cross validation. Root-mean-squared error was implemented as the cost function for training RadSynth. The Haralick GLCM entropy map was constructed using the original equation developed by Haralick et al. [4] and used as the ground truth for training RadSynth.

### 2.4. RadSynth Evaluation and Statistical Analysis

Three different regions of interests (ROIs) of lesion, contralateral glandular, and fatty tissue were segmented from the breast post-contrast DCE-MRI to evaluate the performance of RadSynth. The percentage difference and the Pearson correlation between the entropy values used from the Haralick and RadSynth GLCM entropy map were computed to evaluate the efficacy of radiomic synthesis using RadSynth. The Bland Altman technique was implemented to identify any systematic bias between the Haralick and Radsynth GLCM entropy maps [8]. Statistical significance was set at $p \leq 0.05$.

## 3. RESULTS

The RadSynth model generated synthesized entropy feature maps with excellent accuracy from twenty-four breast patients. The time taken to train the RadSynth model for each cross-validation fold for 50 epochs was approximately two hours. The average time taken by RadSynth to generate entropy images for one patient was 11.8 seconds, significantly lower than traditional methods. **Figure 2** illustrates the corresponding Haralick and RadSynth GLCM entropy maps one from each patient group (normal, benign, and malignant). Radsynth demonstrated excellent correlation with the Haralick GLCM entropy maps as shown in **Table 1,** with an average correlation of 0.97. The mean percentage difference between the entropy values across all the patients and tissue types was $0.07 \pm 0.06$. The scatterplots between the entropy values computed using the two methods corresponding to fatty, glandular, and lesion tissue are illustrated in **Figure 3**.

**Table 1.** Summary of the Pearson correlation coefficient and percentage differences between the entropy values computed by Haralick and RadSynth GLCM.

(a) Correlation coefficient

|  | Fatty tissue | Glandular tissue | Lesion Tissue |
|---|---|---|---|
| Normal Patients | 0.97 | 0.93 |  |
| Benign Patients | 0.97 | 0.97 | 0.94 |
| Malignant Patients | 0.98 | 0.97 | 0.95 |

(b) Percentage difference

|  | Fatty tissue | Glandular tissue | Lesion Tissue |
|---|---|---|---|
| Normal Patients | 0.06±0.04 | 0.06±0.05 |  |
| Benign Patients | 0.07±0.05 | 0.07±0.05 | 0.06±0.08 |
| Malignant Patients | 0.08±0.07 | 0.07±0.05 | 0.07±0.06 |

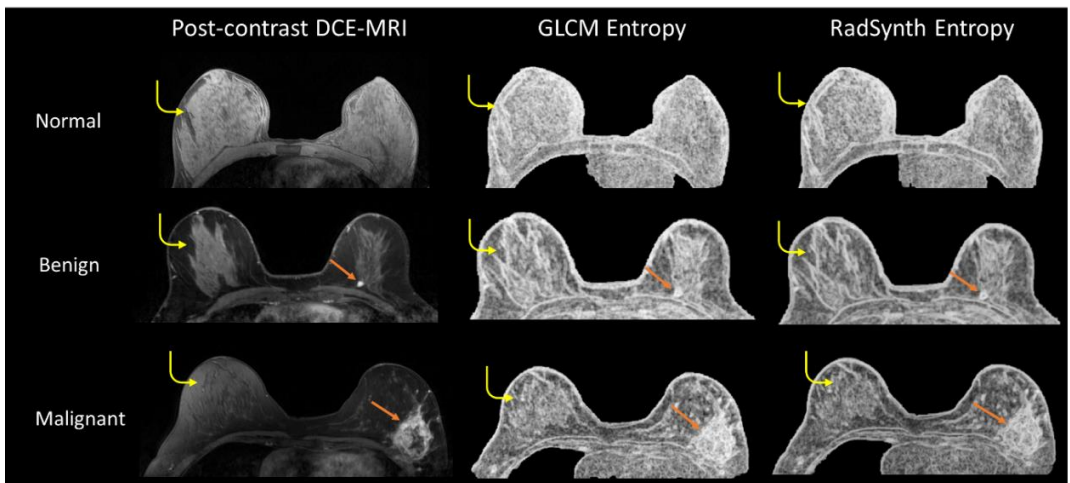

*Figure 2.* Comparison between the GLCM entropy feature maps and RadSynth entropy feature maps for three representative cases, one from each group – normal, benign, and malignant. The lesion is marked using straight orange arrows, while the normal glandular tissue is marked using curved yellow arrows.

## 4. DISCUSSION

RadSynth produced synthetic GLCM entropy images from post-contrast DCE-MRI with excellent accuracy and high correlation. Furthermore, RadSynth took only 11.8 seconds on average to compute radiomic images for each patient, compared to approximately forty minutes using traditional methods. As a result, RadSynth not only reduces the time taken to produce and analyze large radiological datasets in research setting, but also makes it possible to potentially translate radiomic imaging from research to clinical setting.

The major problem with radiomics is the interpretability of radiomic features. For example, there is no specific meaning attached to an entropy value of 5. It could mean that the underlying tissue is heterogeneous or homogeneous depending on the number of gray levels chosen to quantize the radiological image. RadSynth could be used to improve the interpretability of radiomic features. We could use advanced visualization techniques for CNN [9] and decode the textural composition of the underlying tissue in terms of simple components (edges or blobs) or more complex textural representations. Consequently, RadSynth opens up a whole new technique for interpreting radiomic values that did not exist earlier.

There are some limitations to this study. This is a preliminary study to test the efficacy of the newly developed radiomic synthesis mode, RadSynth, in generating radiomic images. This study only tested RadSynth on a single imaging modality for computation of one type of radiomic feature. In the future, extensive validation of RadSynth would be required on datasets with different organs, gray level quantization, imaging modalities and radiomic features.

In conclusion, RadSynth presents a new powerful tool for fast computation of radiomics from radiological images and provides a new perspective into the visualization and analysis of textural information present in the radiological images.

## 5. ACKNOWLEDGEMENTS

National Institutes of Health grant numbers: 5P30CA006973 (IRAT), U01CA140204, and 1R01CA190299 and The Tesla K40 used for this research was donated by the NVIDIA Corporation

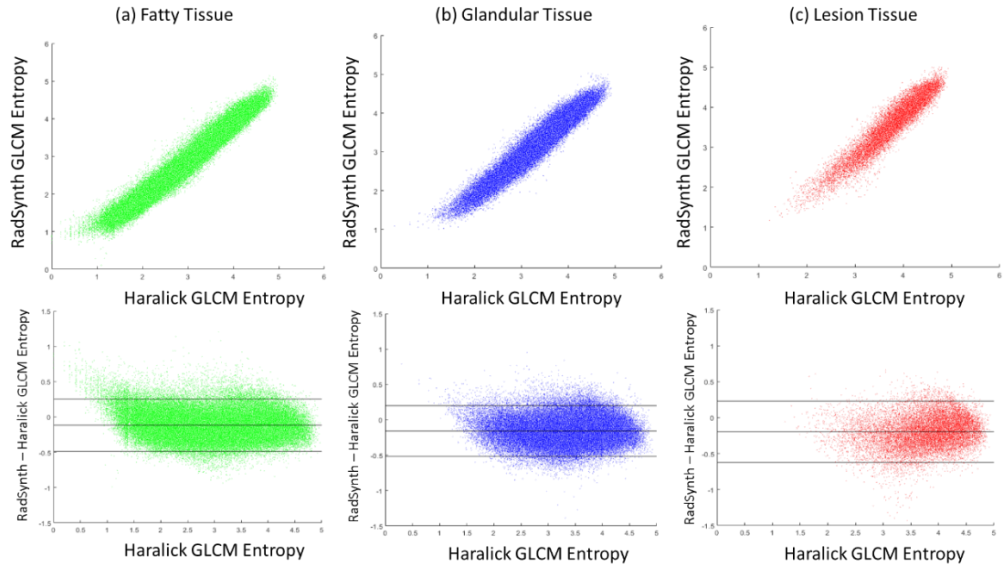

*Figure 3*. Scatterplot (***top row***) and Bland-Altman plot (***bottom row***) between the entropy values computed by RadSynth and Haralick GLCM methods for (a) fatty tissue, (b) glandular tissue, and (c) lesion tissue. There was a high correlation between the entropy values computed by the two methods for each of the three tissue types. ($R_{fat} = 0.98, R_{glandular} = 0.97, R_{lesion} = 0.95$) across all the patients. The Bland-Altman plots demonstrated no bias and excellent agreement between the two datasets.